\documentclass{article}
\usepackage{arxiv}

\usepackage[utf8]{inputenc} % allow utf-8 input
\usepackage[T1]{fontenc}    % use 8-bit T1 fonts
\usepackage{hyperref}       % hyperlinks
\usepackage{url}            % simple URL typesetting
\usepackage{booktabs}       % professional-quality tables
\usepackage{amsfonts}       % blackboard math symbols
\usepackage{nicefrac}       % compact symbols for 1/2, etc.
\usepackage{microtype}      % microtypography
\usepackage[svgnames]{xcolor}
\usepackage{graphicx}
\usepackage{esvect}
\usepackage{amsmath}
\usepackage[tight,footnotesize]{subfigure}
\usepackage{appendix}
\usepackage{lscape}
\usepackage{textcomp}
\usepackage{amssymb}
\usepackage{lscape}
\usepackage[noend]{algpseudocode}
\usepackage[boxruled]{algorithm2e}
\usepackage{hhline}
\usepackage{multirow}
\usepackage{lettrine}
\usepackage{tabularx}

\usepackage{multicol}
\usepackage{mathtools}
\usepackage[english]{babel}
\usepackage{amsthm}
\newtheorem{theorem}{Theorem}[section]
\newtheorem{lemma}[theorem]{Lemma}
\makeatletter
\def\BState{\State\hskip-\ALG@thistlm}
\makeatother

\title{UNFIS: A Novel \underline{N}euro-\underline{F}uzzy \underline{I}nference \underline{S}ystem with \underline{U}nstructured Fuzzy Rules for Classification}

\author{
  Armin Salimi-Badr \\
  Faculty of Computer Science and Engineering\\
  Shahid Beheshti University\\
  Tehran, Iran \\
  \texttt{a\_salimibadr@sbu.ac.ir} \\
}

\begin{document}
\maketitle

\begin{abstract}
An important constraint of Fuzzy Inference Systems (FIS) is their structured rules defined based on evaluating all input variables. Indeed, the length of all fuzzy rules and the number of input variables are equal. However, in many decision-making problems evaluating some conditions on a limited set of input variables is sufficient to decide properly (unstructured rules). Therefore, this constraint limits the performance, generalization, and interpretability of the FIS. To address this issue, this paper presents a neuro-fuzzy inference system for classification applications that can select different sets of input variables for constructing each fuzzy rule. To realize this capability, a new \textit{fuzzy selector neuron} with an adaptive parameter is proposed that can select input variables in the antecedent part of each fuzzy rule. Moreover, in this paper, the consequent part of the Takagi-Sugeno-Kang FIS is also changed properly to consider only the selected set of input variables. To learn the parameters of the proposed architecture, a trust-region-based learning method (\textit{General quasi-Levenberg-Marquardt (GqLM)}) is proposed to minimize cross-entropy in multiclass problems. The performance of the proposed method is compared with some related previous approaches in some real-world classification problems. Based on these comparisons the proposed method has better or very close performance with a parsimonious structure consisting of unstructured fuzzy.\\

\textbf{Keywords:}Fuzzy Neural Networks, Interpretability, quasi-Levenberg-Marquardt (qLM), Unstructured Fuzzy Rules.
\\
\end{abstract}

%% main text
\section{Introduction}
\label{section1}
 \lettrine[findent=2pt]{F}uzzy neural network (FNN) \cite{ANFIS,DENFIS,SOFMLS,PGSOFNN,ICFNN,IT2CFNN,TSKICFNN} is an adaptive and neural form of a rule-based fuzzy system which consists of a set of "IF-THEN" rules. Consequently, it combines the learning capability of a neural network with the interpretability and transparency of a rule-based fuzzy system.

 %Although traditionally an FNN structure is considered as a fuzzy system with "IF-THEN" rules, it is recently shown that it is not valid to explain the output of a fuzzy system based on "IF-THEN" rules, and it is valid to explain it as: "these linguistic antecedents are symptomatic of this output" \cite{Mendel2022}.

An FNN is an universal approximator that realizes a function which maps an input space to the output space (y = f(x)) \cite{Kosko94,Mendel2022}. Therefore, it is proper for different applications that require function approximation, including classification, regression, and control problems. The efficiency of FNNs as a powerful tool is investigated in different applications, including data mining, image processing, system modeling, stock prediction, robotics, and control \cite{Zhang16,Liu16,Wang16,Kim15,fuzz1,fuzz2,fuzz3,fuzz4}.

The well-known adaptive neuro-fuzzy inference system (ANFIS) has been proposed in \cite{ANFIS} that realizes an adaptive neural form of a Takagi-Sugeno-Kang (TSK) fuzzy inference system\cite{Takagi85}. It partitions each input variable into some fuzzy sets uniformly. Afterward, the Gradient Descent method is used to learn the network's parameters. Although the uniform partitioning simplifies the labeling of the linguistic variables, it produces lots of fuzzy rules.

According to the interpretability of the FNNs as "IF-THEN" fuzzy rules, different clustering approaches have been applied to initialize the antecedent along with consequent parts' parameters \cite{CFNN,Teslic11,ICFNN,IT2CFNN}. However, some previous methods, considering the dynamic nature of real-world problems, utilize online structure identification \cite{PGSOFNN,DENFIS,SOFMLS,GPFNN,Salimi-Badr2017,ALMMo0,UNINULL,EFNNNullUni,EFNHN,EFNN}. Generally, a hierarchical online self-organizing learning mechanism is used that adds new fuzzy rules based on the lack of coverage of the current set of fuzzy rules confronting an input sample.

Generally, FNNs form complete and structured fuzzy rules that investigate the values of all input variables in their premise parts. However, checking \textit{all} input variables is not required for decision-making. In many applications checking some conditions on a subset of input variables is sufficient for deciding appropriately. For example in the well-known Fisher Iris problem \cite{IRI}, we can classify different types of flowers by checking their petal length and petal width values. The constraint of using all input variables in FNNs causes forming extra fuzzy rules to check specific linguistic values of a required subset of input variables along with all linguistic values of the other input variables. These extra fuzzy rules decrease the interpretability of the method along with its efficiency due to the \textit{curse of dimensionality}.

Recently, some methods have tried to overcome the problem of extra fuzzy rules by considering the interactions among input variables \cite{CFNN,ICFNN,TSKICFNN,IT2CFNN,Lemos2011,Teslic11,GENEFIS,Panfis}. In \cite{Lemos2011} the \textit{Mahalanobis Distance} is used instead of the \textit{Euclidean Distance} to consider the correlations among input variables. In \cite{CFNN,ICFNN,TSKICFNN,IT2CFNN} non-separable fuzzy rules are constructed by considering the interactions among input variables. In these methods, the input variables are transformed into new non-interactive feature space for constructing each fuzzy rule. Next, the fuzzy sets and rules are defined in these new feature spaces. Although these methods have proper performance in different applications, they violate the interpretability of the FNN by extracting higher-level features from the initial input space. Moreover, these methods are suitable for regression problems and not for classification ones.

In \cite{deepfnn} a subset of input variables are randomly selected. Next, the fuzzy sets are defined based on these selected features. However, the number of variables in the antecedent parts of fuzzy rules remains fixed. Indeed, each fuzzy rule should check linguistic values for all selected features.

Recently some novel methods have tried to solve this problem by introducing FNNs which are able to construct rules based on disjunction (\textit{T-Conorm}) along with conjunction (\textit{T-Norm}) of fuzzy sets for the input variables in their antecedent parts \cite{ALMMo0,UNINULL,EFNNNullUni,DisFuzz,genexp}. In \cite{ANDOR} the idea of using both "AND" and "OR" neurons is proposed. This concept in \cite{ALMMo0,UNINULL,EFNNNullUni,ESODA} is realized by defining a new fuzzy logic neuron (\textit{unineuron}) based on the concept of uninorm. This new neuron can change its function between "AND" and "OR" operators. Therefore, each fuzzy rule is constructed based on "AND" or "OR" connections of the antecedents. In \cite{ESODA} a stochastic approach of a random combination of different logical neurons for binary classification is proposed. In \cite{DisFuzz}, a new layer of "OR" operator is added between the fuzzification layer and the rule layer. Next, the network is applied for regression problems. In \cite{genexp} the gene expression programming is applied to the interval type-2 fuzzy rough neural network to generate fuzzy rules based on various logic operators. These methods have a great performance on real-world applications along with proper interpretability. These structures can create incomplete fuzzy rules because for a problem with \textit{n} input variables, a fuzzy rule that uses "OR" connections can be interpreted as \textit{n} fuzzy rules that check conditions for just one of the input variables. Therefore, these networks can construct \textit{n-ary} and \textit{unary} fuzzy rules. However, these networks can not form \textit{m-ary} ($1 < m < n$) fuzzy rules. In \cite{ESODA,EFNNNullUni}, an \textit{importance weight} is defined for each input variable in each fuzzy rule to show the importance of that feature in firing the fuzzy rule. Although these methods have proper interpretability and good performance, they violate the standard form of \textit{definite} clauses as the form of rules in the knowledge-base which is satisfied in making \textit{Mamdani} and \textit{TSK} Fuzzy Inference Systems.

In this paper, first, to construct an FNN with unstructured fuzzy rules, the logical requirements of relaxing the effect of an input variable on firing a fuzzy rule are investigated. Next, a new fuzzy neuron (\textit{fuzzy selection neuron}) is introduced to select each input variable for defining each fuzzy rule. Therefore, the network can learn to select proper input variables for defining each fuzzy rule.

For adjusting different parameters, a learning algorithm is required. Different learning methods including \textit{Stochastic Gradient Descent} (SGD), Linear Least Square (LLS), and Levenberg-Marquardt (LM) methods have been applied for learning different parameters of FNNs. The LLS is a popular method that has two constraints: 1- It is applicable for linear output neurons which are not proper for classification problems, and 2- It can adjust only the consequent parts' parameters of the fuzzy rules. Moreover, SGD can be utilized for training the premise parts' parameters based on \textit{error backpropagation}, but it has the problem of \textit{vanishing gradient} due to the number of layers and form of activation functions in an FNN. One of the most successful learning methods applied for FNNs is LM \cite{CFNN,ICFNN,IT2CFNN}. The main idea behind the LM is to utilize the linear approximation of the Mean-Squared-Error (MSE) based on the \textit{Taylor's expansion} while satisfying a \textit{trust region}. The final algorithm performs like a second-order optimization method (e.g. \textit{Newton-Raphson's} method) that approximates the \textit{Hessian matrix} based on the multiplication of \textit{Jacobian matrix} with its transposed \cite{Menhaj94,CFNN,ICFNN,Rubio21,IT2CFNN}. Although the loss function used in the LM method is proper for regression problems,
it is not appropriate for classification. Instead, the \textit{Cross-Entropy} is a suitable loss function for classification problems \cite{goodfellow2016deep}.

In this paper, a trust-region-based learning algorithm similar to the LM is presented to minimize the Cross-Entropy for a multiclass classification problem. We call the novel learning algorithm the General quasi-Levenberg-Marquardt (GqLM). The proposed learning method approximates the \textit{Hessian matrix} of the loss function based on the quadratic form of the \textit{Jacobian matrix}. We can summarize the contributions of this paper as follows:
\begin{enumerate}
\item Proposing a new fuzzy neuron (fuzzy selection neuron) to select different input variables for defining each fuzzy rule based on logical characteristics of \textit{T-Norm} operators;
\item Presenting a novel Fuzzy Neural Network (UNFIS) that is able to construct incomplete and unstructured rules based on the proposed fuzzy selection neuron;
\item Presenting a trust-region learning method inspired by the Levenberg-Marquardt (LM) for multiclass classification problems (GqLM);
\item Adjusting the parameters of the proposed UNFIS by applying the proposed GqLM learning method.
\end{enumerate}

The rest of this paper is organized as follows. In Section \ref{section2}, the structure of the concept of unstructured rules, the proposed fuzzy selection neuron, the proposed architecture, and the proposed learning method are presented and discussed in detail. Next, section \ref{section3} reports the effectiveness of the proposed method in different real-world classification problems. Finally, conclusions are presented in Section \ref{section4}.

\section{Proposed Method}
\label{section2}
In this section first, the structure of fuzzy rules is described and the proposed feature selection approach to remove the effect of some input variables and form unstructured rules is presented. Afterward, the proposed architecture of the Fuzzy Neural Network with the new fuzzy selection neuron is explained. Next, the General quasi-Levenberg-Marquardt (GqLM) learning method is presented.

\subsection{Unstructured Fuzzy Rules}
\label{UFR}
The $i^{th}$ fuzzy rule of a Multi-Input-Multi-Output (MIMO) Takagi-Sugeno-Kang (TSK) Fuzzy Inference System (FIS) can be expressed as follows \cite{Takagi85,SUGENO198815,mendel2017uncertain}:
\begin{equation}
\small
\mathcal{R}_i: \bigwedge_{j = 1}^n \left(x_j \, is\, A_j^i\right)\, \Rightarrow \, \bigwedge_{c = 1}^C\left(y^i_c = f(\alpha_{c,0}^i + \sum_{j=1}^n\alpha_{c,j}^i.x_j)\right)
    \label{eq:stfr}
\end{equation}
where $\wedge$ is the symbol of logical conjunction (\textit{T-Norm} operator), $x_j$ is the $j^{th}$ input variable, $n$ is the number of input variables, $y_c^i$ is the proposed $c^{th}$ output based on the $i^{th}$ fuzzy rule, $f$ is a nonlinear function, $C$ is the number of outputs, and $\alpha_{c,j}^i$s are consequent part's parameters of the $i^{th}$ fuzzy rule for $c^{th}$ output.

Although such a standard and complete fuzzy rule takes into account all the input variables, in some cases the decision making process does not require to investigate conditions on all input variables. In these cases only a subset of input variables is required. Indeed the length of such fuzzy rules (number of antecedent fuzzy sets and considered input variables) could be between 1 and $n$. We can define such incomplete and unstructured fuzzy rule as follows:
\begin{equation}
\small
\mathcal{UR}_i: \bigwedge_{j \in \mathcal{D}^i} \left(x_j \, is\, A_j^i\right)\, \Rightarrow \, \bigwedge_{c = 1}^C\left(y^i_c = f(\alpha_{c,0}^i + \sum_{j \in \mathcal{D}^i}\alpha_{c,j}^i.x_j)\right)
    \label{eq:unfr}
\end{equation}
where $\mathcal{D}^i$ is a set containing indices of the considered subset of input variables in the $i^{th}$ fuzzy rule. Indeed, its main difference with the former format is the limited number of input variables used for its definition. This proposed unstructured rule is the general TSK form of the \textit{Incomplete Zadeh Fuzzy Rule} defined in \cite{mendel2017uncertain}.

Such unstructured incomplete fuzzy rule can be put into the format of the standard complete one by treating the unconsidered input variables as the member of fuzzy set $\mathcal{T}$ (named after \textit{tautology}) that is always \textit{True} for all values of an input variable as follows \cite{mendel2017uncertain}:
\begin{equation}
\small
\mathcal{UR}_i: \bigwedge_{j = 1}^n \left(x_j \, is\, A_j^{+i}\right)\, \Rightarrow \, \bigwedge_{c = 1}^C\left(y^i_c = f(\alpha_{c,0}^i + \sum_{j=1}^n\alpha_{c,j}^{+i}.x_j)\right)
    \label{eq:ufr2}
\end{equation}
where $A^+$ and $\alpha^{+}$ are fuzzy set and consequent parameter that consider the selection of an input variable in a fuzzy rule, respectively and can be defined as follows:
\begin{equation}
    A^{+i}_j = \left\{\begin{array}{ll}
    A^{i}_j; & j \in \mathcal{D}^i \\
    \mathcal{T}; & j \not\in \mathcal{D}^i
    \end{array}\right.
\label{eq:ufs}
\end{equation}

\begin{equation}
    \alpha^{+i}_j = \left\{\begin{array}{ll}
    \alpha^i_j; & j \in \mathcal{D}^i \\
    0; & j \not\in \mathcal{D}^i
    \end{array}\right.
\label{eq:augalpha}
\end{equation}

In an adaptive FIS, there is no a prior knowledge to determine the set of considered input variables of different fuzzy rules ($\mathcal{D}^i$) and these sets should be derived in a learning process. Moreover, the length of all fuzzy rules are not the same. Therefore, we need an adaptive parameter for selecting proper input fuzzy sets. Moreover, it is required to have a correlation between selecting an input variable in the antecedent part of a rule and considering its effect on the consequent part of the same rule. Furthermore, it is better to have a fuzzy selection process. In the other words, an input variable can take part in a fuzzy rule partially.
\begin{figure}[t]
\centering
\includegraphics[width = 3.5 in]{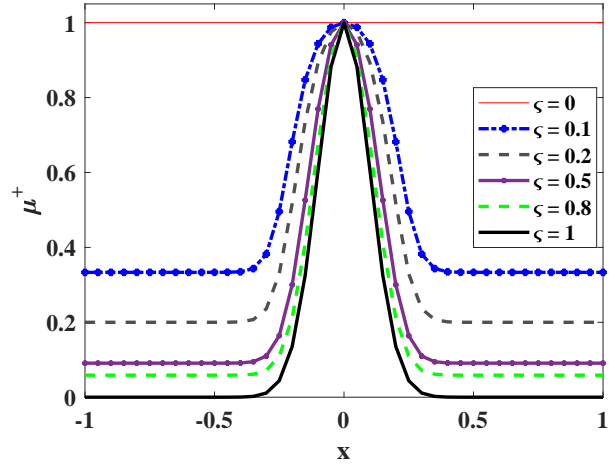}\caption{The effect of changing value of selection parameter $\varsigma$ on the shape of a Gaussian membership function.}
\label{fig:differentshapes}
\end{figure}
To overcome the mentioned problems we define the selection parameter $\varsigma_j^i$ for selecting the $j^{th}$ input variable in the $i^{th}$ fuzzy rule as follows:

\begin{equation}
    \varsigma_j^i = \frac{1}{1+exp(-s_j^i)}
    \label{eq:varsigma}
\end{equation}
where $s_j^i$ is an adaptive parameter which its proper value would be determined in the learning process. Based on using the \textit{Logistic map} in eq. (\ref{eq:varsigma}), $\varsigma_j^i$ is in range (0,1).

If we define two fuzzy sets $A_j^i$ and $A_j^{+i}$ for the $j^{th}$ input variable with universe $X_j$ in $i^{th}$ fuzzy rule as follows:

\begin{equation}
    A_j^i = \int_{x_j \in X_j} \mu_j^i/x_j
    \label{eq:A}
\end{equation}

\begin{equation}
    A_j^{+i} = \int_{x_j \in X_j} \mu_j^{+i}/x_j
    \label{eq:Aplus}
\end{equation}

To realize the definitions presented in eqs. (\ref{eq:ufs}) and (\ref{eq:augalpha}) with an adaptive form the following equations are proposed:
\begin{equation}
    \mu_j^{+i} = \frac{\mu_j^i+\epsilon}{(1-\varsigma_j^i).\mu_j^i + \varsigma_j^i+\epsilon}
    \label{eq:muplus}
\end{equation}
\begin{equation}
    \alpha_j^{+i} = \varsigma_j^i.\alpha_j^i
    \label{eq:alphaplus}
\end{equation}
where $0 < \epsilon << 1$ is a very small positive constant that is added to nominator along with the denominator to have a membership value greater than zero. These equations can realize the required selecting process in eqs. (\ref{eq:ufr2}) to (\ref{eq:augalpha}) based on the value of $\varsigma_j^i$ as follows:
\begin{equation}
\begin{array}{ll}
 \left\{\begin{array}{l}
    \lim_{\varsigma_j^i \to 1}\mu^{+i}_j \approx \mu_j^i \\
    \lim_{\varsigma_j^i \to 0}\mu^{+i}_j = 1 \\
    \end{array}\right.
    ;&  \left\{\begin{array}{l}
    \lim_{\varsigma_j^i \to 1}\alpha^{+i}_j = \alpha_j^i \\
    \lim_{\varsigma_j^i \to 0}\alpha^{+i}_j = 0 \\
    \end{array}\right.
\end{array}
    \label{eq:lim}
\end{equation}
which means that if $\varsigma_j^i$ is close to 1, the $j^{th}$ variable would be considered in both antecedent and consequent part of the $i^{th}$ rule (selected); while if this parameter is close to zero, the effect of this input variable is relaxed (not be selected). Fig. \ref{fig:differentshapes} shows the effect of changing selecting parameter $\varsigma$ on the membership value for a Gaussian membership function.

The following Lemmas show that the proposed formula in eq. (\ref{eq:muplus}) has the characteristics of a fuzzy set similar to the initial fuzzy set $A$.

\begin{lemma}
The membership value $\mu^{+i}_j$ is in range (0,1).
\label{lemma1}
\end{lemma}

\begin{proof}
Since both numerator and denominator of $\mu^{+i}_j$ in eq. (\ref{eq:muplus}) are positive value, $\mu^{+i}_j$ is positive ($\mu^{+i}_j > 0$). To show that its value is lower than 1, we assume that its value is greater than 1 knowing that $0 < \mu_j^i < 1$ and $0 < \varsigma_j^i < 1$:
\begin{equation}
\begin{array}{lll}
     &\mu_j^{+i} = \frac{\mu_j^i+\epsilon}{(1-\varsigma_j^i).\mu_j^i + \varsigma_j^i +\epsilon} > 1 & \\
     \Rightarrow & \mu_j^{i} > (1-\varsigma_j^i).\mu_j^i + \varsigma_j^i &\\
     \Rightarrow & \varsigma_j^i.\mu_j^i > \varsigma_j^i \Rightarrow  \mu_j^i > 1 \quad \scalebox{1.5} {\textbf{\textreferencemark}}
\end{array}
\end{equation}
\end{proof}

\begin{lemma}
The centers of fuzzy sets $A^{+i}_j$ and $A^{i}_j$ are equivalent.
\label{lemma2}
\end{lemma}

\begin{proof}
If $m_j^i$ is the center of $A^{i}_j$, it has the maximum membership value ($\mu_j^{i}/m_j^i = 1$ and $\frac{d\mu_j^i}{dx}|_{x = m_j^i} = 0$). From Lemma \ref{lemma1} we know that $\mu_j^{+i} \in (0,1)$. For $m_j^i$ we have:
\begin{equation}
\begin{array}{lll}
     &\mu_j^{i}/m_j^i = 1 &\\
     \Rightarrow & \mu_j^{+i}/m_j^i = \frac{\mu_j^i/m_j^i+\epsilon}{(1-\varsigma_j^i).(\mu_j^i/m_j^i) + \varsigma_j^i+\epsilon} &= \frac{1+\epsilon}{(1-\varsigma_j^i) + \varsigma_j^i+\epsilon}=1\\\\
     & \frac{d\mu_j^{+i}}{dx} = \frac{\varsigma_j^i}{\left((1-\varsigma_j^i).(\mu_j^i) + \varsigma_j^i+\epsilon\right)^2}.\frac{d\mu_j^{i}}{dx}&\\
     \Rightarrow & \frac{d\mu_j^{+i}}{dx}|_{x = m_j^i} = \frac{d\mu_j^i}{dx}|_{x = m_j^i} = 0 &
\end{array}
\end{equation}
Therefore, $m_j^i$ has the maximum value of $\mu_j^{+1}$ and it is the center of $A^{+i}_j$.
\end{proof}
From Lemma \ref{lemma2} we can conclude that $A^{+i}_j$ preserves the linguistic value of $A^i_j$.
\subsection{Unstructured Neuro-Fuzzy Inference System (UNFIS)}
\label{UNFIS}

\begin{figure}[t]
\centering
\includegraphics[width = 3.5 in]{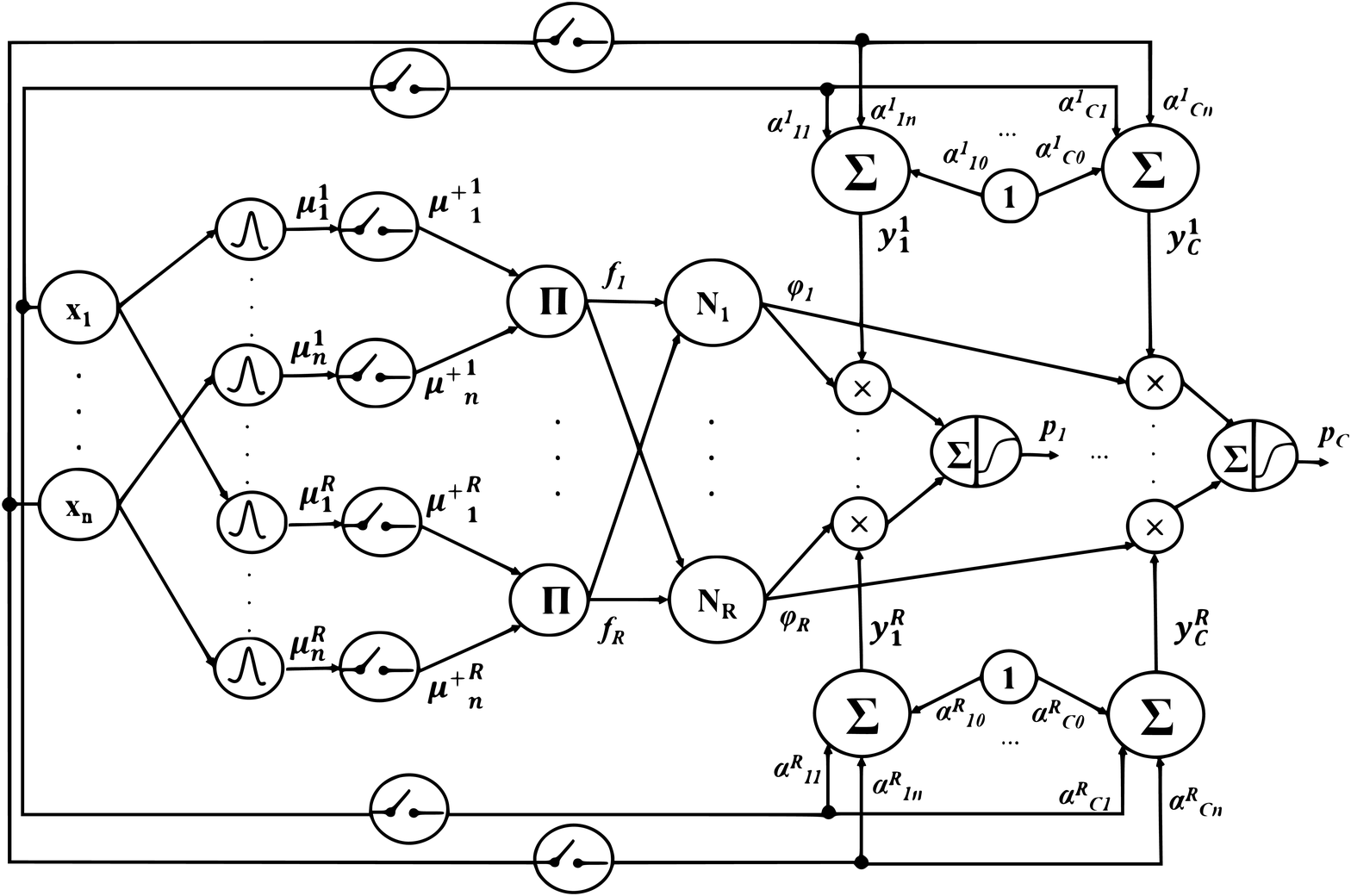}
\caption{The architecture of the proposed Fuzzy Neural Network.}
\label{fig:unfis}
\end{figure}

The structure of the proposed fuzzy neural network (UNFIS) is shown in Fig. \ref{fig:unfis}. This structure realizes a TSK adaptive neuro-fuzzy inference system for classification tasks, which is able to select the effect of different input variables in constructing different fuzzy rules. Consequently, it can form unstructured fuzzy rules with different length (considering the effect of different number of input variables). The main difference between the proposed structure and a standard TSK-FNN is the \textit{Fuzzy Selection Neurons} that select different input variables and are shown in Fig. \ref{fig:unfis} as switches.

Totally, the proposed network is composed of seven distinct layers:

\begin{enumerate}
\item \textbf{Input Layer}: The first layer is the input layer that provides the network with the input vector including the input variables. Its output is input instance as follows:
\begin{equation}
X = [x_1,  x_2,  \cdots, x_n]^T
\end{equation}

\item \textbf{Fuzzification Layer}: The fuzzy neurons of this layer plays the role of fuzzy sets and calculates the Gaussian membership value of each input variable $x_j$ ($j =1,2, \cdots, n$) to a fuzzy set of a fuzzy rule ($A^i_j$, $i=1,2, \cdots, R$) as follows:

\begin{equation}
\mu_j^i = exp\left(-\frac{(x_j-m_j^i)^2}{2.(\sigma_j^i)^2}\right)
\label{eq:fs}
\end{equation}
where, $m_j^i$ and $\sigma_j^i$ are the center and width of the corresponding Gaussian fuzzy set for $j^{th}$ input variable in the $i^{th}$ fuzzy rule.

\item \textbf{Fuzzy Selection Layer}: This layer consists of \textit{Fuzzy Selection Neurons} (shown in Fig. \ref{fig:unfis} as switches) to decide whether the value of an input variable should be taken into account for a fuzzy rule or not. The output of these neurons for $j^{th}$ input variable and the $i^{th}$ rule is $\mu^{+i}_j$ which is calculated based on eq. (\ref{eq:varsigma}) and eq. (\ref{eq:muplus}).

\item \textbf{Fuzzy Rules Layer}:
The neurons of this layers computes the firing strength of each fuzzy rule based on applying the fuzzy conjunction (\textit{T-Norm}) on the membership values of the selected input variables. Here, the \textit{dot-product} is utilized as the \textit{T-Norm} operator. The firing strength of the $i^{th}$ fuzzy rule is calculated as follows:

\begin{equation}
f_i = \Pi_{j=1}^n\mu_j^{+i}
\end{equation}

\item \textbf{Normalization Layer}: The calculated firing strength values of the previous layer is normalized by the neurons of this layer as follows:

\begin{equation}
\phi_i = \frac{f_i}{\sum_{l=1}^R f_l}
\end{equation}

\item \textbf{Consequent Layer}: In this layer, the proposed output of each fuzzy rule for each class ($Y^i_c$) is calculated based on the consequent part's parameters ($\alpha_{c,j}^{+i}$ for $i^{th}$ fuzzy rule, $j^{th}$ input variable, and $c^{th}$ class that is previously introduced in eq. (\ref{eq:alphaplus})) and the computed normalized firing strength ($\phi_i$) as follows:

\begin{equation}
\left\{
\begin{array}{l}
y^i_c = \alpha_{c,0}^i + \sum_{j=1}^n\alpha_{c,j}^{+i}.x_j \\
Y^i_c = \phi_i.y^i_c
\end{array}\right.
\label{eq:conseq}
\end{equation}
where $\alpha^{+i}_j$ is calculated based on eq. (\ref{eq:alphaplus}).

\item \textbf{Output Layer}: The final output of the network for an input vector is computed based on the \textit{one-hot encoding} in this layer. Indeed, the $c^{th}$ neuron of this layer represents the probability of classifying the input vector as an instance of $c^{th}$ class. Therefore, the number of neurons in this layer is equal to the number of classes ($C$). The activation function of these neurons is \textit{Soft-Max} and the output of $c^{th}$ neuron is calculated as follows:

\begin{equation}
p_c = \frac{\sum_{i=1}^R (Y_c^i)-\theta_c)}{\sum_{g=1}^C (\sum_{i=1}^R (Y_g^i)-\theta_g)}
\label{eq:output}
\end{equation}
where $\theta_c$ is a threshold value for $c^{th}$ class.

\end{enumerate}

\subsection{General quasi-Levenberg-Marquardt (GqLM) Algorithm}

We assume that the vector $\pi$ contains all adaptive parameters of the network and its prediction for categorizing the $k^{th}$ training instance $X_k$ in $c^{th}$ class is denoted as function $F_c(X_k;\pi)$. If we change the parameter vector $\pi$ as $\pi_{new} = \pi + \Delta \pi $ in the learning process satisfying a trust region, we can approximate the new network's output linearly based on the \textit{Taylor's expansion} as follows:

\begin{equation}
    \begin{aligned}
    \left\{\begin{array}{lll}
         \hat{P}_c^{new} &= &\left[\begin{array}{c}
             F(X_1;\pi+\Delta \pi),
             \cdots,
             F(X_N;\pi+\Delta \pi)
        \end{array}
        \right]^T\\
        \hat{P}_c &= &\left[\begin{array}{c}
             F(X_1;\pi),
             \cdots,
             F(X_N;\pi)
        \end{array}
        \right]^T \\
        \hat{P}_c^{new}  &=& \hat{P}_c + J_c\times\Delta \pi
    \end{array}
    \right.
    \end{aligned}
    \label{eq_lm1}
\end{equation}
where, $\hat{P}_c$ and $\hat{P}^{new}_c$ are network's prediction for $c^{th}$ class with parameters $\pi$ and $\pi+\Delta \pi$, and $J_c$ is the \textit{Jacobian matrix} for $c^{th}$ class defined as follows:

\begin{equation}
    J_c = \frac{\partial \hat{P}_c}{\partial \pi} =     \left[\begin{array}{ccc}
        \frac{\partial \hat{p}_{c,1}}{\partial \pi_1} & \cdots & \frac{\partial \hat{p}_{c,1}}{\partial \pi_{n_\pi}} \\
        \vdots &  \ddots & \vdots \\
        \frac{\partial \hat{p}_{c,N}}{\partial \pi_1} & \cdots & \frac{\partial \hat{p}_{c,N}}{\partial \pi_{n_\pi}}
    \end{array}\right]
    \label{eq_lm2}
\end{equation}
Here, $\hat{p}_{c,k}$ (c = 1,2, $\cdots$, C  $\&$ k = 1,2, $\cdots$, N) is the $c^{th}$ output of the network for $k^{th}$ training instance (the estimated probability of categorizing the $k^{th}$ instance as a member of the $c^{th}$ class), $n_{\pi}$ is the number of parameters included in the vector $\pi$, and $\pi_i$ (i=1, ..., $n_\pi$) is the $i^{th}$ adjustable parameter.

Now we define the loss function $\mathcal{L}$ as follows:

\begin{equation}
\small
    \mathcal{L} = \sum_{c=1}^C\left(-P_c^{*T} log(\hat{P}_c^{new}) + \eta.\left(\Delta \pi^T J_c^T J_c \Delta \pi + \lambda.\Delta \pi^T \Delta \pi\right)\right)
    \label{eq_lm4}
\end{equation}
where, $P^*_c$ is a vector contains the desired outputs (labels) of training instances for the $c^{th}$ class, $T$ is the symbol of \textit{transpose} matrix, $\eta$ and $\lambda$ are constant positive penalty coefficients. This proposed loss function is composed of three parts: 1- the cross-entropy for a multi-class classification problem, 2- a regularizing term to limit the amount of change on network's output ($\Delta \pi^T J_c^T J_c \Delta \pi = \Vert J_c \Delta \pi\rVert_2^2 )$, and 3- a regularization term to limit the amount of change on the network' parameters ($\Delta \pi^T \Delta \pi = \Vert\Delta \pi\rVert_2^2 )$). These penalty terms define the trust region and realize the applied \textit{Taylor's expansion} in eq. (\ref{eq_lm1}). Indeed, considering these penalty terms avoid applying dramatic changes on the parameters and network's outputs.

Since the amount of changing the parameter vector is penalized in eq. (\ref{eq_lm4}), we can linearly approximate the logarithmic part of the loss function based on the \textit{Taylor's expansion}. Moreover, we substitute the linear approximation of $P^{new}_c$ presented in eq. (\ref{eq_lm1}). Consequently, the loss function can be rewritten as follows:

\begin{equation}
\small
\begin{array}{ll}
     \mathcal{L} & = -\sum_{c=1}^C\left(P_c^{*T}\left(\log(\hat{P}_c)+\Xi_c.J_c.\Delta \pi\right)\right)  \\
     &+      \eta.\sum_{c=1}^C\left(\left(\Delta \pi^T (J_c^T J_c + \lambda.I)\Delta \pi\right)\right)
\end{array}
    \label{eq_lm5}
\end{equation}
where, $I$ is the identity matrix, and $\Xi$ is a diagonal matrix that contains the inverse of network's outputs as follows:

\begin{equation}
    \Xi_c = \left[\begin{array}{cccc}
        \frac{1} {\hat{p}_{c1}}& 0  & \cdots & 0 \\
        0 & \frac{1} {\hat{p}_{c2}} & \cdots & 0 \\

        \vdots &  \vdots & \ddots & \vdots\\
        0 & 0 & \cdots & \frac{1} {\hat{p}_{cN}}
    \end{array}\right]
    \label{eq_lm6}
\end{equation}

Based on the optimality necessary condition \cite{Boyd}, the optimal change parameter vector $\Delta \pi$ is derived by solving the following equation:

\begin{equation}
\begin{array}{lll}
     &\frac{\partial \mathcal{L}}{\partial \Delta \pi} &= 0  \\
     \Rightarrow & \Delta \pi & = \frac{1}{\eta}J^+\left(\sum_{c=1}^C\left(J^T_c \Xi_c P^*_c\right)\right)
\end{array}
    \label{eq_lm7}
\end{equation}
where, $J^+$ is an auxiliary variable computed as follows:

\begin{equation}
\begin{array}{lll}
J^+ = (\sum_{c=1}^C\left(J^T_c.J_c+\lambda.I\right))^{-1}
\end{array}
    \label{eq_lm8}
\end{equation}

To complete the learning algorithm and update the parameter vector $\pi$ based on eq. (\ref{eq_lm7}), we should derive the matrices $J_c$s presented in eq. (\ref{eq_lm2}). The set of adaptive parameters of the proposed network consists of centers and width values of fuzzy sets ($m_j^i$ and $\sigma_j^i$ in eq. (\ref{eq:fs})), selection parameters ($s_j^i$ in eq. (\ref{eq:varsigma})), consequent parts parameters ($\alpha_{c,j}^i$ in eq. (\ref{eq:conseq})), and the threshold values ($\theta_c$ in eq. (\ref{eq:output})). The total number of parameters is equal to $(3+C)Rn+(R+1)C$, where $R$ is the number of fuzzy rules, $C$ is the number of classes and $n$ is the number of input variables. The form of parameter vector $\pi$ is as follows:
\begin{equation}\small
 \begin{array}{cccccccccc}
     \pi =& [m_1^1& m_2^1& \cdots & m_n^1 &
     m_1^2 & \cdots & m_n^R & &\sigma_1^1\\ &\cdots & \sigma_n^R & s_1^1 & \cdots & s_n^R & \alpha_{1,0}^1 &
     \alpha_{1,1}^1 & \cdots & \alpha_{1,n}^1\\& \alpha_{2,0}^1& \cdots& \alpha_{C,n}^1& \alpha_{1,0}^2 &
     \cdots & \alpha_{C,n}^R& \theta_1& \cdots &  \theta_C]^T
\end{array}
    \label{parameter vector}
\end{equation}
To compute \textit{Jacobian matrix} defined in eq. (\ref{eq_lm2}) the following derivatives derived based on using the \textit{Chain rule} should be calculated for each training sample ($X = [x_0, x_1, x_2 , ..., x_n]^T$ and $x_0 = 1$):

\begin{equation}
    \frac{\partial \hat{p}_{c}}{\partial \alpha_{gj}^i} =         (\delta_{(c,g)}.\hat{p}_{c}.(1-\hat{p}_{c})-(1-\delta_{(c,g)})\hat{p}_{c}.\hat{p}_{g}).\phi_i.x_j.\varsigma_j^i
    \label{eq:j1}
\end{equation}

\begin{equation}
\begin{array}{ll}
     \frac{\partial \hat{p}_{c}}{\partial s_{j}^i} & = \sum_{g=1}^C((\delta_{(c,g)}.\hat{p}_{c}.(1-\hat{p}_{c})\\
     &-(1-\delta_{(c,g)})\hat{p}_{c}.\hat{p}_{g}).\sum_{r=1}^R(y^r_g.(\delta_{(r,i)}.\phi_i.(1-\phi_i)\\
     &+(1-\delta_{(r,i)})\phi_i.\phi_r.f_i).\frac{\varsigma_j^i.(1-\varsigma_j^i).(1-\mu_j^i)}{(1-\varsigma_j^i)\mu_j^i+\varsigma_j^i})\\
     &+\phi_i.x_j.\alpha_{g,j}^i.\varsigma_j^i.(1-\varsigma_j^i))
\end{array}
    \label{eq:j2}
\end{equation}

\begin{equation}
\begin{array}{ll}
     \frac{\partial \hat{p}_{c}}{\partial m_{j}^i} & = \sum_{g=1}^C((\delta_{(c,g)}.\hat{p}_{c}.(1-\hat{p}_{c})\\
     &-(1-\delta_{(c,g)})\hat{p}_{c}.\hat{p}_{g}).\sum_{r=1}^R(y^r_g.(\delta_{(r,i)}.\phi_i.(1-\phi_i)\\
     &+(1-\delta_{(r,i)})\phi_i.\phi_r.f_i).\frac{\varsigma_j^i}{(1-\varsigma_j^i)\mu_j^i+\varsigma_j^i}.\frac{(x_j-m^i_j)}{\sigma_j^{i2}}))\\
\end{array}
    \label{eq:j3}
\end{equation}

\begin{equation}
\begin{array}{ll}
     \frac{\partial \hat{p}_{c}}{\partial \sigma_{j}^i} & = \sum_{g=1}^C((\delta_{(c,g)}.\hat{p}_{c}.(1-\hat{p}_{c})\\
     &-(1-\delta_{(c,g)})\hat{p}_{c}.\hat{p}_{g}).\sum_{r=1}^R(y^r_g.(\delta_{(r,i)}.\phi_i.(1-\phi_i)\\
     &+(1-\delta_{(r,i)})\phi_i.\phi_r.f_i).\frac{\varsigma_j^i}{(1-\varsigma_j^i)\mu_j^i+\varsigma_j^i}.\frac{(x_j-m^i_j)^2}{\sigma_j^{i3}}))\\
\end{array}
    \label{eq:j4}
\end{equation}

\begin{equation}
     \frac{\partial \hat{p}_{c}}{\partial \theta_{g}} = -\left((\delta_{(c,g)}.\hat{p}_{c}.(1-\hat{p}_{c})
     -(1-\delta_{(c,g)})\hat{p}_{c}.\hat{p}_{g})\right)
    \label{eq:j5}
\end{equation}
where, for two different variables $x$ and $y$, $\delta_{x,y}$ is the \textit{Kronecker delta function} that returns 1 if $x=y$, and 0 otherwise. Considering that $\epsilon$ is very small and negligible, the effect of this parameter is neglected in the above derivations. Further details of the learning algorithm are presented in Algorithm \ref{alg1}.

\begin{algorithm}[t]
\caption{General quasi-Levenberg-Marquardt}\label{alg1}
\hspace*{\algorithmicindent} \textbf{Input:} \\
\hspace*{\algorithmicindent}$\quad\,$
Training samples $X$ along with their desired outputs $P_c^*$\\
\hspace*{\algorithmicindent}$\quad\,$
Number of classes ($C$), rules ($R$), and samples ($N$) \\
\hspace*{\algorithmicindent}$\quad\,$ Mini-batch size ($M$)\\
 \hspace*{\algorithmicindent} $\quad$ \text{Learning rates $\lambda$, $\beta$ and $\eta$} \\
 \hspace*{\algorithmicindent} $\quad$ \text{Maximum number of iterations $iter_{max}$}\\
 \hspace*{\algorithmicindent} $\quad$ \text{Initialized parameters ($m_{j}^i$, $\sigma_j^i$, $s^i_j$, $\alpha_{c,j}^i$, and $\theta_c$)}\\
\hspace*{\algorithmicindent} \textbf{Output:}\\
\hspace*{\algorithmicindent}$\quad$ \text{Learned parameters ($m_{j}^i$, $\sigma_j^i$, $s^i_j$, $\alpha_{c,j}^i$, and $\theta_c$)}\\
 \text{$\Delta \pi^* \gets \Vec{0}$;}\\
\text{Form the parameter vector $\pi$ (eq. (\ref{parameter vector}));}\\
\For {$iter \gets 1$ \textbf{to} $iter_{max}$}{
\text{Shuffle training samples $X$;}\\
\text{$index \gets 1$;}\\
\While{index $\leq$ N}{
\For {$k \gets index$ \textbf{to} $index + M$}{
\text{Calculate the network's outputs:}\\
\text{$\quad \hat{P}_c, \phi_i, f_i, y^i_c, \varsigma_j^i$}\\
\text{$\quad$ for $k^{th}$ instance (eq. (\ref{eq:varsigma}), and eqs.(\ref{eq:fs}) to (\ref{eq:output}));}\\
\For {$c \gets 1$ \textbf{to} $C$}{
\text{Complete the $k^{th}$ row of $J_c$ (eq. (\ref{eq_lm2}), and eqs. (\ref{parameter vector}) to (\ref{eq:j5}));}\\
\text{Complete the $k^{th}$ row of $\Xi_c$ (eq. (\ref{eq_lm6}), and eq. (\ref{eq:output});}\\
}
}
\text{Compute $J^+$ (eq. (\ref{eq_lm8}));}\\
\text{Compute $\Delta \pi$ (eq. (\ref{eq_lm7}));}\\
\text{$\Delta \pi^* \gets \beta.\Delta \pi^* + (1-\beta).\Delta \pi$;}\\
\text{$\pi \gets \pi + \Delta \pi^*$;}\\
\text{Update parameters;}\\
\text{$index \gets index + M$;}\\
}
}
\end{algorithm}

\section{Experiments}
\label{section3}
In this section, the performance and interpretability of the proposed network in real-world classification problems are investigated. First, performance of the model is compared with the performance of some related methods in classifying some binary and multi-class real-world problems. Afterward, the ability of the proposed model to form unstructured fuzzy rules with different number of antecedents is studied for a well-known classification problem. Finally, the effect of the proposed learning method (GqLM) is studied.

\subsection{Experimental Setup and Performance Criteria}

To evaluate the model, Accuracy (ACC) and Area Under the Receiver Operating Characteristics (ROC) Curve (AUC) are used as follows \cite{EFNNNullUni,UNINULL}:

\begin{equation}
    ACC = \frac{TP+TN}{TP+TN+FN+FP}\times 100
\end{equation}
\begin{equation}
    AUC = 0.5\left(\frac{TP}{TP+FN}+\frac{TN}{TN+FP}\right)
\end{equation}
where, $TP$ is the number of positive instances classified correctly, $FP$ is the number of negative instances classified incorrectly, $TN$ is the number of negative instances classified correctly, and $FN$ is the number of positive instances classified incorrectly. To follow previous studies \cite{EFNNNullUni,UNINULL}, for binary classification problems, both measures are utilized, while for the multiclass ones, only the Accuracy is used.

To show the effect of the proposed fuzzy selection neuron, we define the number of active features for $i^{th}$ fuzzy rule as follows:
\begin{equation}
n_{af}^i = \sum_{j=1}^n\varsigma_j^i
    \label{eq:AF}
\end{equation}

Table \ref{table_param} shows the values used for the hyperparameters of the model for all problems. These values are chosen based on try-and-error. Generally, users can determine the proper values for these parameters based on different search methods.

\begin{table}[t!]\scriptsize
    \caption[c]{Hyperparameters of the proposed algorithm.}
    \begin{center}
    \begin{tabular}{|c|c|c|c|c|c|}
        \hline
         Minibatch Size (M)& $\lambda$ & $\eta$ & $\beta$ & $Iter_{max}$  & Number of Rules (R)\\
         \hline
         32 & 1e3 & 1e-3 & 0.9 & 100 & 2 \\
         \hline
    \end{tabular}
    \end{center}
    \label{table_param}
\end{table}

Moreover, to initialize the antecedent and consequent parts parameters, a centroid-based clustering method that utilizes K-Nearest-Neighbors is applied \cite{IT2CFNN}.

The performance of the proposed method is compared with the most related models that construct fuzzy rules based on using both conjunction and disjunction operators as follows:
\begin{itemize}
    \item \textbf{EFNN-NullUni} \cite{EFNNNullUni}: An evolving FNN that constructs fuzzy rules based on using the \textit{null-unineuron} (both \textit{AND} and \textit{OR} operators) and weighting the input variables;
    \item \textbf{UNInull-FNN} \cite{UNINULL}: An evolving FNN that utilize 2-uninorms to aggregate the fuzzified input variables;
    \item \textbf{ALMMo-0*} \cite{ALMMo0}: a model to autonomously zero-order multiple learning with pre-processing that improves the classifier’s accuracy, as it creates stable models;
    \item \textbf{EFNHN} \cite{EFNHN}: This network utilizes artificial hydrocarbon network for defuzzification and combines an evolving fuzzification technique, training based on extreme learning machine, and automatic relevance determination;
    \item \textbf{EFNN} \cite{EFNN}: An evolving model that uses unineurons in constructing fuzzy rules and apply pruning based on automatic relevance determination;
    \item \textbf{FNN}: A normal TSK neuro-fuzzy system that constructs structured fuzzy rules and learned with the proposed GqLM method. Indeed, this network is very similar to the proposed network but it does not utilize the proposed \textit{fuzzy selection neuron} and has not the second layer of UNFIS. Therefore, its performance is compared with UNFIS to show the effect of unstructured rules. Its parameters' values, initialization method, and the applied fine-tuning algorithm are same as those of UNFIS.
\end{itemize}

All experiments have been conducted in MATLAB 2018b runtime environment on an Intel Core-i3 CPU with clock speed 2.1 GHz and 8 GB RAM running the Windows 10 professional operating system.

\subsection{Real-world Classification Problems}

In these experiments, the performance of the proposed model in classifying 6 real-world benchmark problems is compared with some relevant methods. The number of input variables of these problems is various (from 3 to 13) and they include binary and multiclass problems. Their input variables are majorly high-level and expert-understandable. Table \ref{table_1} summarizes the characteristics of these benchmark problems. To follow previous studies (\cite{EFNNNullUni,UNINULL}), the algorithm is executed 30 times and in each execution, 70$\%$ of the samples are randomly selected as the training dataset and the rest for test.

Table \ref{table_4} shows the number of active features selected for constructing each fuzzy rule for each problem based on eq. (\ref{eq:AF}), and Tables \ref{table_2} and \ref{table_3} compare the performance of model on test data, with some relevant previous methods. Based on the reported results, majorly  the proposed method has the best performance. Moreover, based on these results it is obvious that the performance of the model outperforms the other method especially in problems with higher input variables (for example the accuracy is significantly better than the other methods for binary problem \textit{Autism} with 18 input variables and the multiclass \textit{Wine} dataset with 13 input variables). Its accuracy is the second best just for \textit{Habermann} dataset with 3 input variables. Furthermore, in all experiments the performance of the proposed method is significantly better than FNN that constructs fuzzy rules based on all input variables. These observations show the effectiveness of the unstructured scheme with various number of selected features for each fuzzy rule. To show the effectiveness of the proposed adaptive feature selection.

\begin{table}[t!]\small
    \caption[c]{Characteristics of datasets used in the experiments.}
    \begin{center}
    \begin{tabular}{|c|c|c|c|c|c|}
        \hline
         Dataset & Acronym & Number of & Size of & Size of  & class \\
         &  & features & training set & test set &\\
         \hline
         Haberman \cite{HAB} & HAB & 3 & 214&92 & 2\\
         Cryotherapy \cite{CRI} & CRI & 6& 63& 27&2 \\
        Heart \cite{HEA} & HEA&13&189&81 &2\\
        Autism \cite{thabtah2017autism} & AUT&18&737&316 &2\\
        Iris \cite{IRI} &IRI &4&105&45& 3\\
        Thyroid \cite{THY}& THY&5&150&65& 3\\
        Wine \cite{WIN} &WIN&13&124&54 &3\\
         \hline
    \end{tabular}
    \end{center}
 \label{table_1}
 \end{table}

\begin{table}[t]
    \caption[c]{Number of active features in each fuzzy rule for different problems.}
    \begin{center}
    \begin{tabular}{|c|c|c|c|c|c|c|c|}
        \hline
         Dataset & HAB & CRI & HEA &AUT& IRI  &THY & WIN \\
         \hline
         Fuzzy rule 1 &1.33 & 2.55 & 6.77&8.37& 2.29 & 2.95& 7.50  \\
         \hline
         Fuzzy rule 2 & 1.35 & 2.40 & 6.76 & 8.76 &3.31 & 3.23 & 6.64 \\
         \hline
    \end{tabular}
    \end{center}
 \label{table_4}
 \end{table}

\begin{table*}[t!]
\small
    \caption[c]{Comparison of the performance of the proposed method with some relevant models based on Accuracy. The best value is represented in bold text.}
    \begin{center}
    \begin{tabular}{|c|c|c|c|c|c|c|c|}
        \hline
         Dataset & EFNN-NullUni$^\dag$ & UNInull-FNN$^\dag$ & ALMMo-0*$^\dag$& EFNHN$^\dag$ & EFNN$^\dag$ & FNN & UNFIS\\
         \hline
         HAB & \textbf{76.06$\pm$2.99} & 73.36$\pm$3.92 & 71.25$\pm$2.45 & 58.07$\pm$7.37&73.11$\pm$4.32 &74.60$\pm$4.52 &74.20$\pm$4.30 \\
         CRI & 81.26$\pm$7.16 & 82.46$\pm$9.06& 78.25$\pm$1.65 & 70.98$\pm$10.17 & 81.48$\pm$9.72 & 52.47$\pm$10.70&\textbf{84.69$\pm$8.68}\\
         HEA & 80.12$\pm$4.28& 66.17$\pm$10.08&71.12$\pm$6.21&63.25$\pm$5.79&79.14$\pm$4.36&52.19$\pm$3.49&\textbf{81.26$\pm$3.77}\\
         AUT & 95.98$\pm$4.42& 76.37$\pm$2.58&79.99$\pm$1.14&77.61$\pm$8.75&81.36$\pm$4.52&68.71$\pm$0.78&\textbf{98.03$\pm$1.97}\\
         IRI & 96.17$\pm$5.44& 94.51$\pm$3.53&85.98$\pm$7.12&88.59$\pm$12.87&82.96$\pm$9.02&72.11$\pm$17.25&\textbf{96.25$\pm$2.69}\\
         THY&89.54$\pm$3.14&85.07$\pm$ 4.27&81.85$\pm$3.30&65.20$\pm$13.08&82.46$\pm$6.18&79.43$\pm$3.94&\textbf{94.72$\pm$2.70}\\
         WIN & 61.13$\pm$7.43 & 65.34$\pm$14.21 & 61.11$\pm$2.75 &62.66$\pm$7.21 & 62.07$\pm$12.12 &47.32$\pm$10.36& \textbf{74.20$\pm$12.48} \\
         \hline
         \multicolumn{7}{c}{$^\dag$: Results of these methods are adapted from \cite{EFNNNullUni} }\\
    \end{tabular}
    \end{center}
 \label{table_2}
 \end{table*}

\begin{table*}[t!]
\small
    \caption[c]{Comparison of the performance of the proposed method with some relevant models in binary classification problems, based on Area Under the ROC Curve (AUC). The best value is represented in bold text.}
    \begin{center}
    \begin{tabular}{|c|c|c|c|c|c|c|c|}
        \hline
         Dataset & EFNN-NullUni$^\dag$ & UNInull-FNN$^\dag$ & ALMMo-0*$^\dag$& EFNHN$^\dag$ & EFNN$^\dag$ & FNN&UNFIS\\
         \hline
         HAB & 0.5625$\pm$0.02& 0.5755$\pm$0.03& 0.5355$\pm$ 0.07&0.5449$\pm$0.06&0.5435$\pm$0.04&0.5356$\pm$0.0438 &\textbf{0.5793$\pm$0.0515} \\
         CRI & 0.7998$\pm$0.08& 0.8242$\pm$0.08& 0.7855$\pm$ 0.07&0.7013$\pm$0.09&0.7878$\pm$0.07 &0.5483$\pm$0.0720 &\textbf{0.8508$\pm$0.0883}\\
         HEA & 0.7976$\pm$0.03& 0.6439$\pm$0.11&0.6239$\pm$0.09&0.5847$\pm$0.06&0.7743$\pm$0.04&0.4880$\pm$0.0201&\textbf{0.8095$\pm$0.0383}\\
         AUT & 0.9676 $\pm$0.04& 0.7278 $\pm$0.13&0.7562 $\pm$0.06&0.7768 $\pm$0.08&0.8167 $\pm$0.04&0.4998$\pm$0.0046&\textbf{0.9774$\pm$0.1402}\\
         \hline
         \multicolumn{7}{c}{$^\dag$: Results of these methods are adapted from \cite{EFNNNullUni} }\\
    \end{tabular}
    \end{center}
 \label{table_3}
 \end{table*}

\subsection{Illustrative Example for Proposed Feature Selection}
\begin{figure}[t!]
\centering
\begin{tabular}{cc}
    \subfigure[][]{\includegraphics[width = 1.8in]{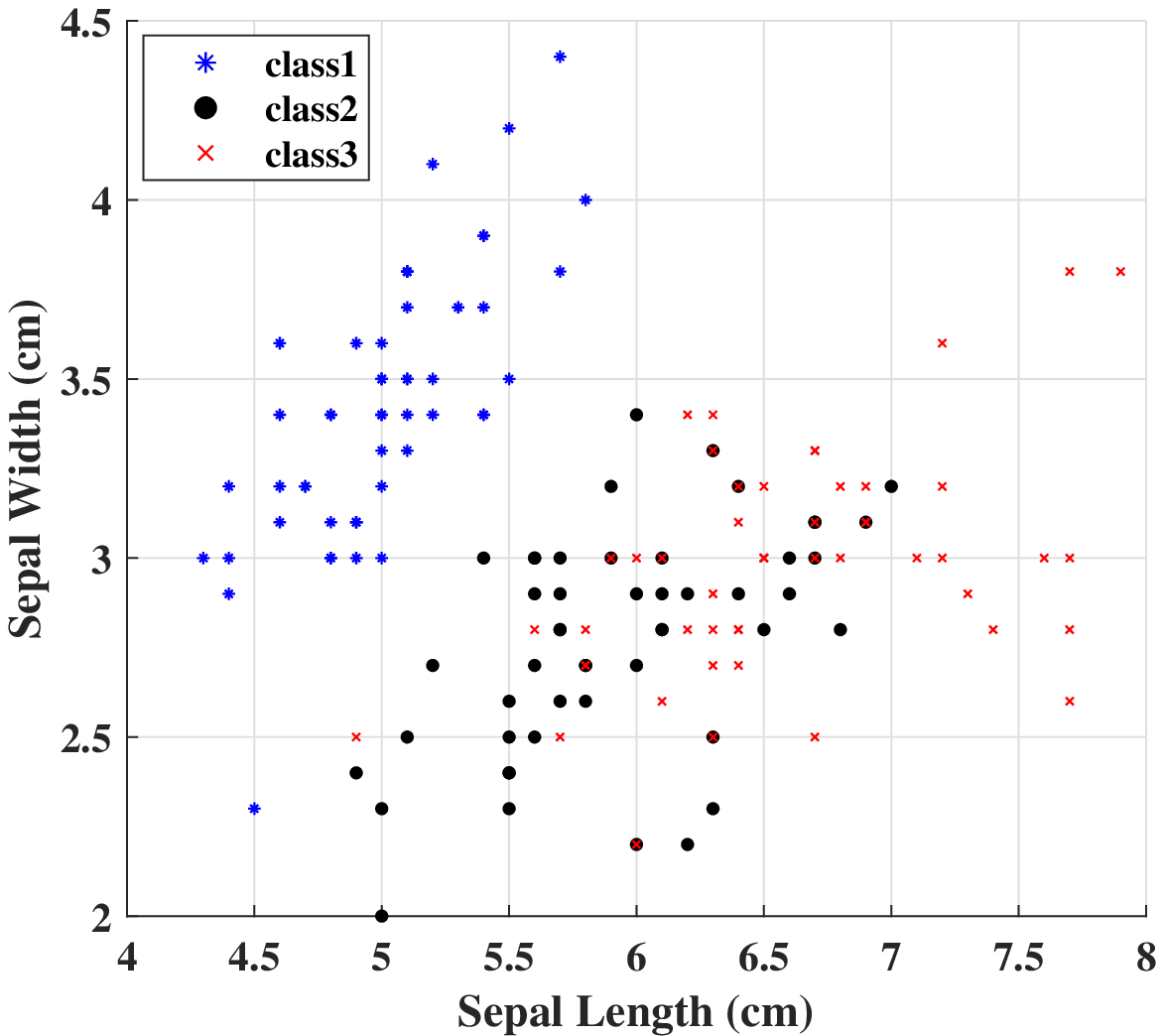}} & \subfigure[][]{\includegraphics[width = 1.8in]{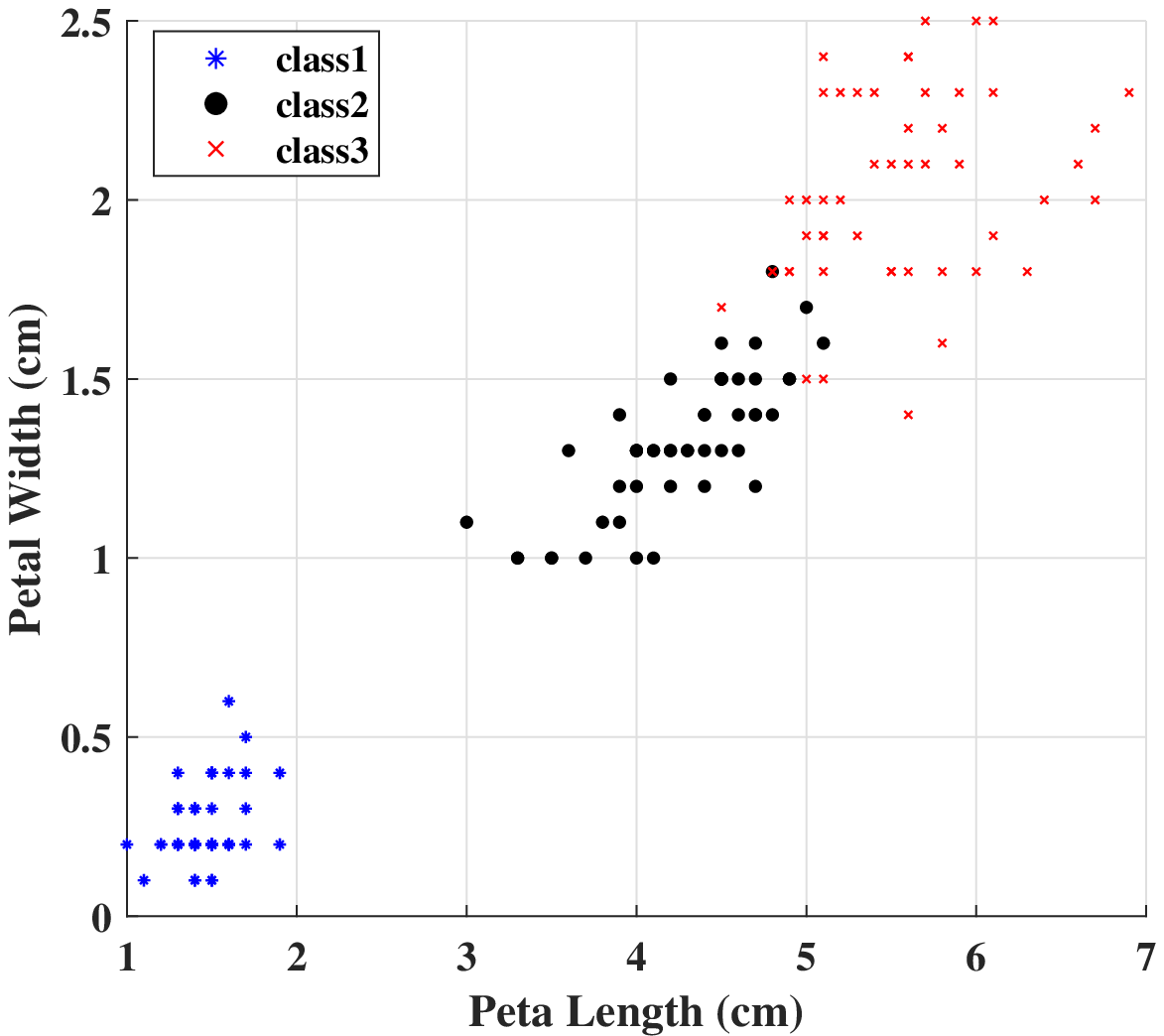}}\\
    \subfigure[][]{\includegraphics[width = 1.8in]{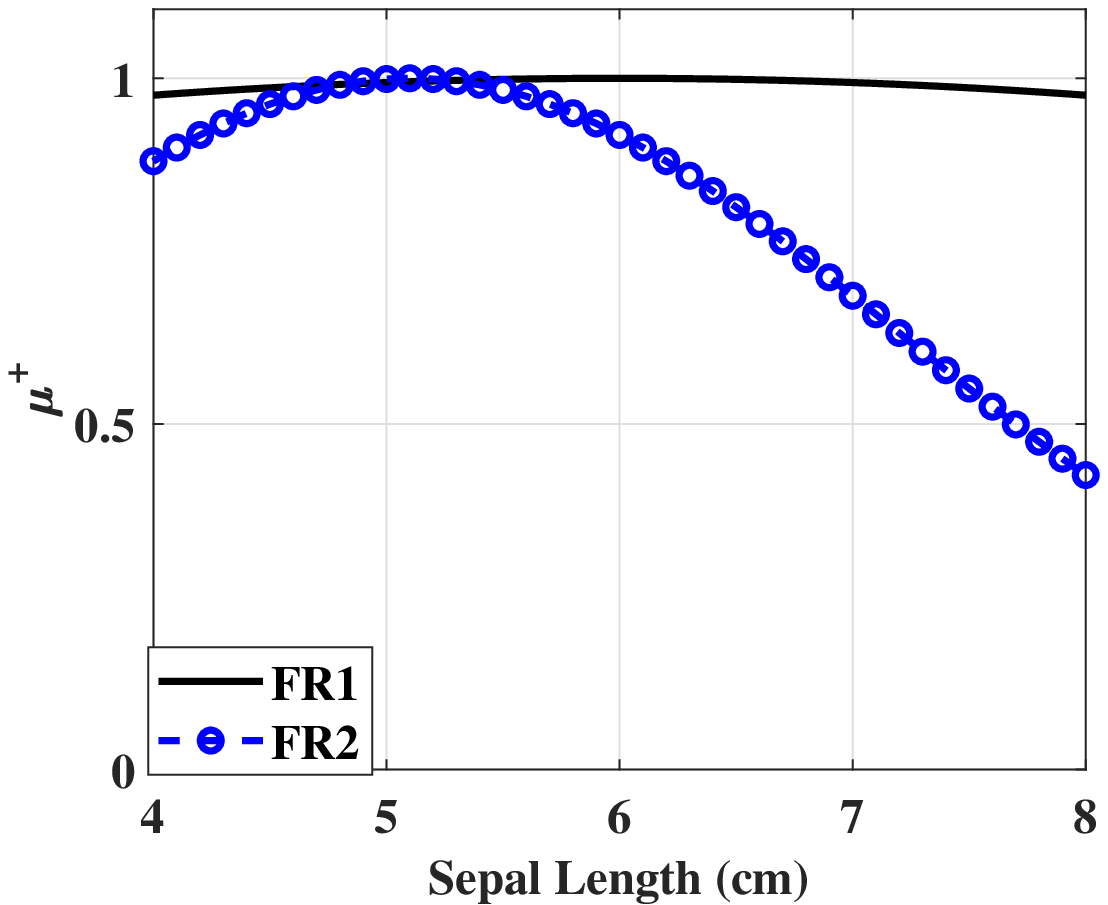}} &
    \subfigure[][]{\includegraphics[width = 1.8in]{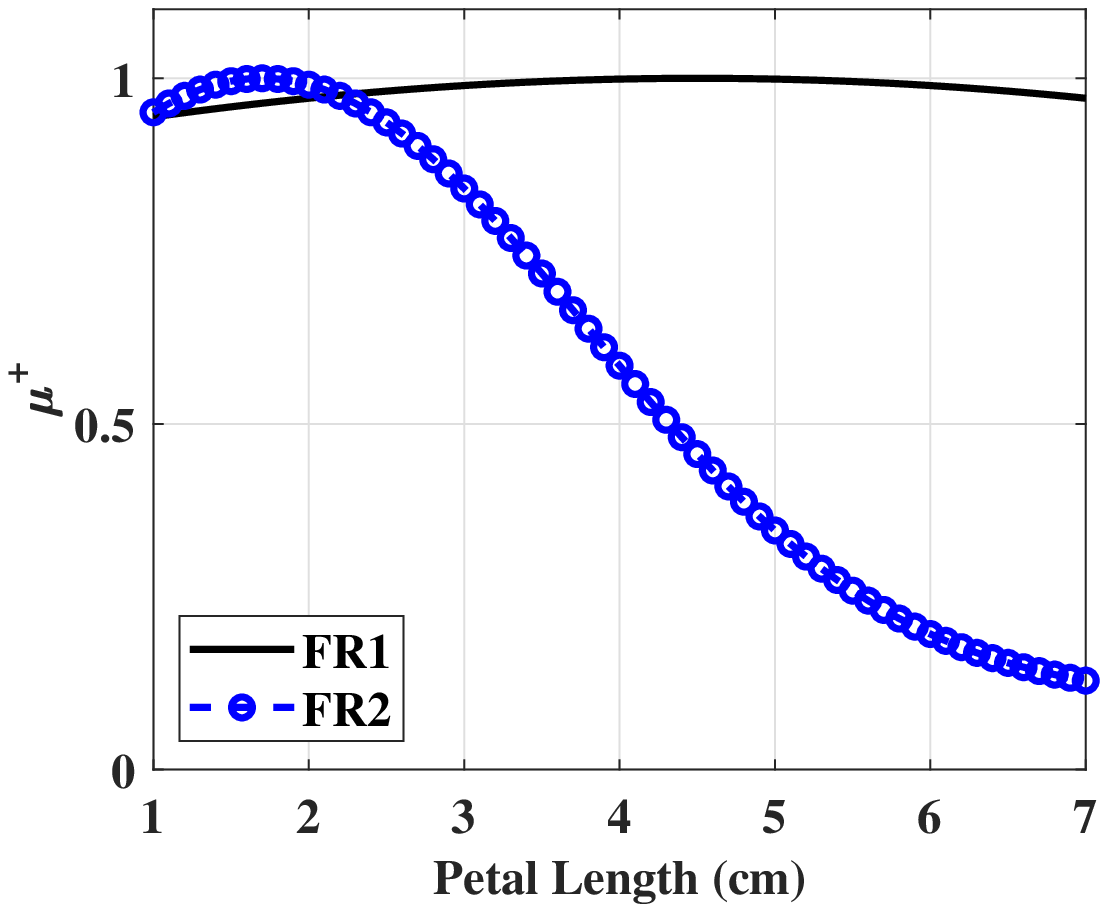}} \\
    \subfigure[][]{\includegraphics[width = 1.8in]{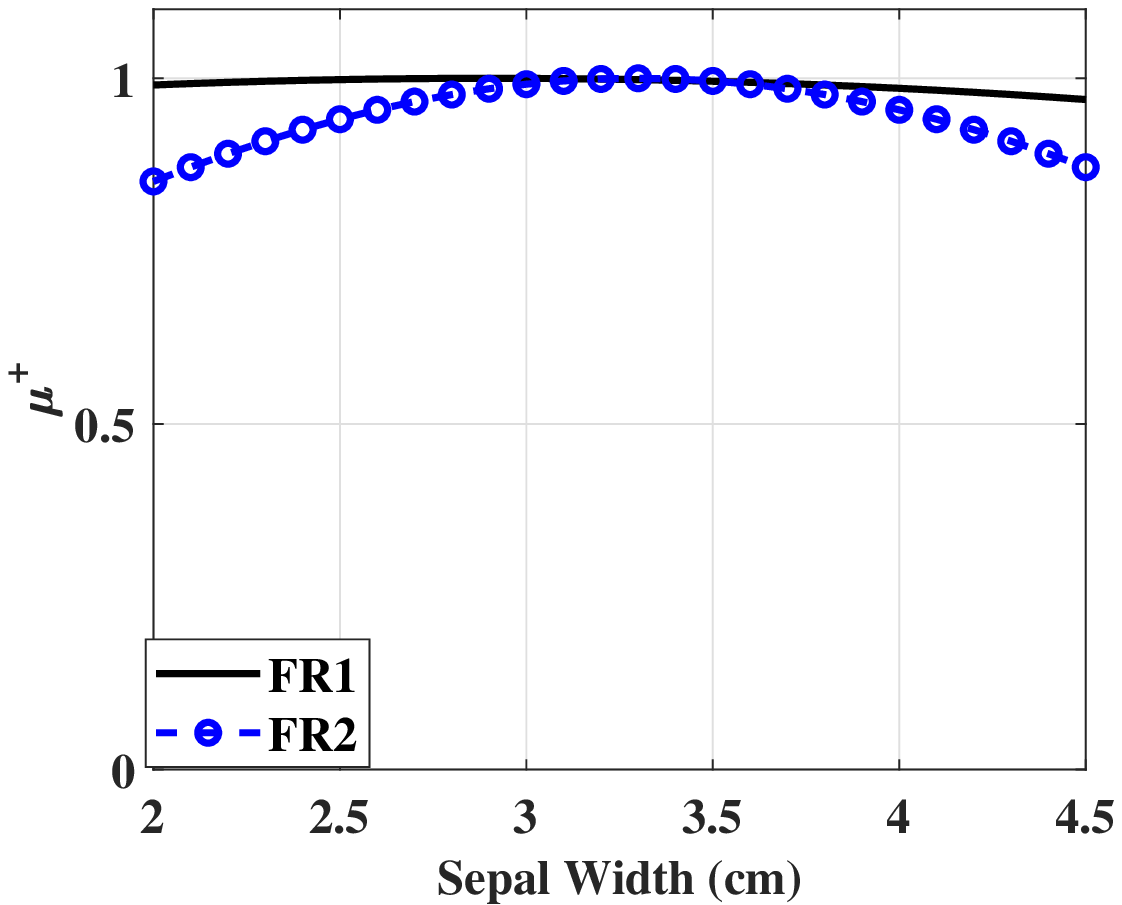}} &
    \subfigure[][]{\includegraphics[width = 1.8in]{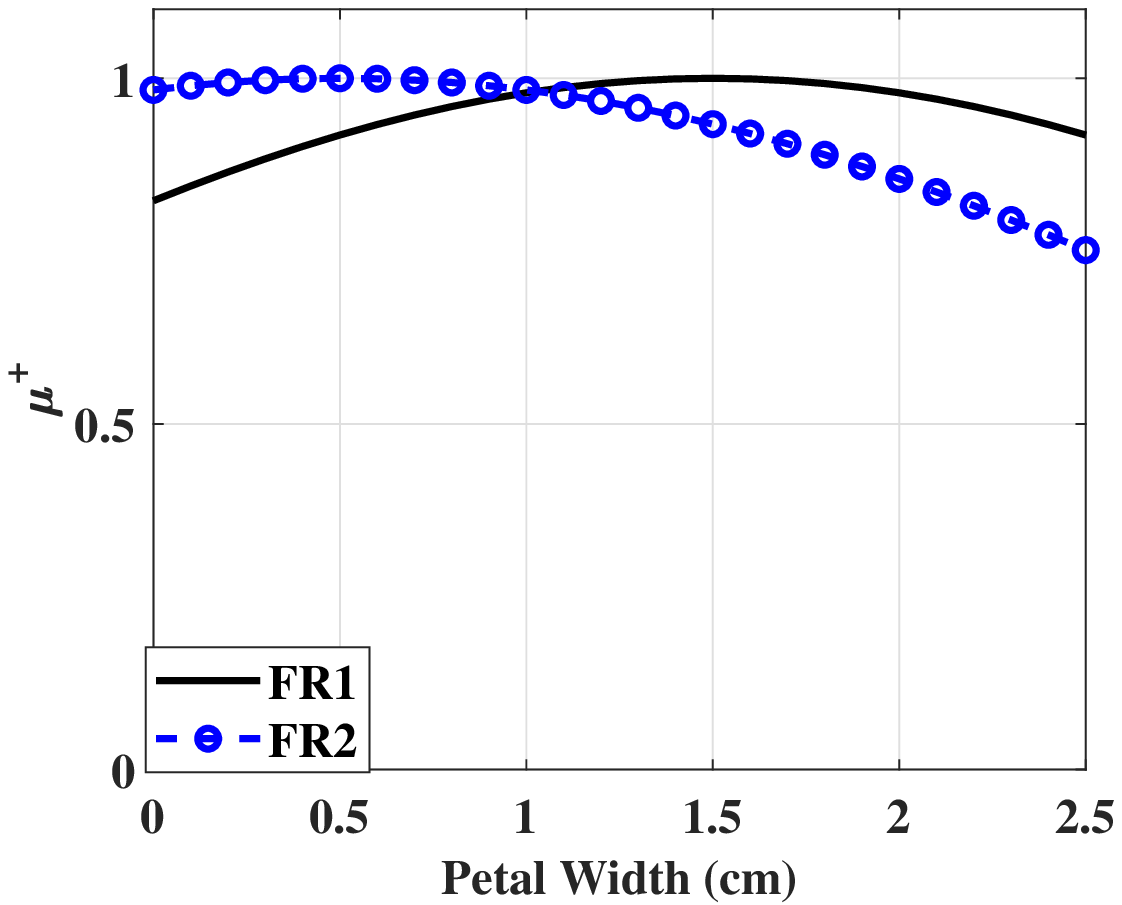}}
\end{tabular}
\caption{Samples of Iris dataset along with the extracted fuzzy sets for two fuzzy rules.}
\label{fig:iris}
\end{figure}

Here, to show the effect of the proposed fuzzy feature selection method in constructing unstructured rules, the well-known Fisher Iris classification problem is utilized. This dataset contains information of three types of iris flowers. It has four input variables including length and width values of sepal and petal of the Iris flowers. Based on Figs. \ref{fig:iris}-(a) and \ref{fig:iris}-(b), values of a subset of input variables (for example petal length and width values) is sufficient to accurately classify the samples of this dataset.

Fig. \ref{fig:iris} shows the extracted fuzzy sets of two fuzzy rules for different input variables of the Iris dataset. It is obvious that the first fuzzy rule (black solid lines) considers mainly the fourth variable (Petal Width) because the membership values of the other input variables are very close to 1. However, the other rule (blue dashed lines) considers all input variables. Moreover, by comparing the centers of extracted fuzzy sets, it is obvious that the goal of the first rule is to decide to categorize samples into class 2 or class 3. On the other hand, the main purpose of the second rule is to distinguish the samples of the first class from the other ones.

\subsection{Studying the Performance of GqLM}
To study the effect of the proposed learning method (GqLM), the  performance of the UNFIS learned by GqLM is compared with its performance learned by some other related methods including Levenberg-Marquardt (LM), Stochastic Gradient (SGD), and Momentum.

\begin{figure}[t]
\centering
\includegraphics[width = 3.25 in]{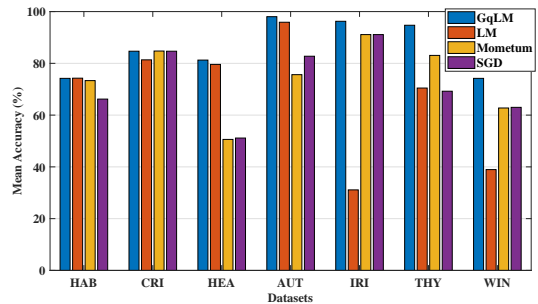}\caption{Comparison of the performance of the proposed learning algorithm (GqLM) with other methods including Levenberg-Marquardt (LM), Stochastic Gradient Descent (SGD), and Momentum.}
\label{fig:compar}
\end{figure}

Fig. \ref{fig:compar} shows the performance of the model after learning by different algorithms in different classification problems. Based on these results, it is shown that the performance of GqLM is better than the other methods in all problems. Its performance is significantly better than SGD and Momentum in the most cases. However, the LM that minimizes the Root Mean Squared Error (RMSE) instead of the Cross-Entropy has good performance in binary problems, but its performance decreases dramatically in multiclass problems (see the performance of LM for Iris and Wine datasets). These results show the importance of using the cross-entropy in multiclass problems.

According to these results, the proposed learning method has proper performance in different problems, especially in multiclass ones.

\section{Conclusions}
\label{section4}
In this paper a neuro-fuzzy inference system able to construct unstructured fuzzy rules (UNFIS) is proposed. The proposed architecture can relax the effect of some input variables in the antecedent parts of fuzzy rules. Therefore, for an input space with $n$ dimensions, the fuzzy rules would be \textit{m-ary} where $1 \leq m \leq n$. Indeed, the network can cancel the effect of redundant input variables.

A new fuzzy neuron named \textit{fuzzy selection neuron} is introduced to relax the effect of the redundant input variables in forming each fuzzy rule. The new neuron can cancel the effect of the initial fuzzy set by increasing the membership value to 1 for the universe of discourse. This neuron has an adaptive parameter to select each input variable for each fuzzy rule.

Furthermore, in this paper, a multiclass trust region learning method inspired by the Levenberg-Marquardt (LM) method, named General quasi-Levenberg-Marquardt (GqLM) is proposed for minimizing the Cross-Entropy. Afterward, the proposed learning method has been successfully applied for learning the premise and consequent parts' parameters of UNFIS.

The performance of the model is evaluated for classifying some binary and multiclass classification problems. Its performance is compared with the most related recently presented Fuzzy Neural Networks that majorly have the ability of constructing rules based on both logical "AND" and "OR". In most cases the proposed method outperforms the other methods.

Moreover, the performance of the proposed learning algorithm (GqLM) is compared with some related methods including Levenberg-Marquardt (LM), Stochastic Gradient Descent (SGD), and Momentum. The proposed method has the best performance in all cases, especially in multiclass problems.

The future trend of this study includes: 1- Expanding the method to a type-2 fuzzy neural network based on \textit{uncertain selection}, 2- Utilizing the Fuzzy Selection Neuron in a Deep Fuzzy Neural Network for solving problems based on raw data like images, and 3- Expanding the architecture to consider interactions among selected input variables and forming unstructured non-separable fuzzy rules.
%\clearpage

\bibliographystyle{ieeetr}
\bibliography{ref}

\end{document}